\documentclass{article} % For LaTeX2e
\usepackage{iclr2027_conference,times}

\usepackage{amsmath,amsfonts,bm}

\def\eqref#1{equation~\ref{#1}}
\def\1{\bm{1}}

\def\mA{{\bm{A}}}

\def\mH{{\bm{H}}}

\def\mK{{\bm{K}}}

\def\mM{{\bm{M}}}

\def\mP{{\bm{P}}}
\def\mQ{{\bm{Q}}}

\def\mZ{{\bm{Z}}}

\DeclareMathAlphabet{\mathsfit}{\encodingdefault}{\sfdefault}{m}{sl}
\SetMathAlphabet{\mathsfit}{bold}{\encodingdefault}{\sfdefault}{bx}{n}

\usepackage{hyperref}
\usepackage{url}
\usepackage{graphicx}
\usepackage{booktabs}
\usepackage{multirow}
\usepackage{colortbl} % Required for \rowcolor
\usepackage{multicol} % Enable multi-column environment
\usepackage{tabulary}
\usepackage{subcaption}
\usepackage{amsmath} % Recommended for mathematical typesetting
\usepackage{amssymb} % Required for \lesssim
\usepackage[noend]{algpseudocode} % noend option for cleaner look
\usepackage{algorithm}
\usepackage{tabularx}   % For creating tables with auto-wrapping text
\DeclareCaptionSubType{algorithm} % enables subalgorithm
\usepackage[most]{tcolorbox} % Load the tcolorbox package
\usepackage{wrapfig,lipsum}
\title{Region-Level Policy Optimization for Fine-grained MLLM Perception}

\author{Yuheng Shi$^{1}$, Xiaohuan Pei$^{1}$, Minjing Dong$^{2}$, Chang Xu$^{1}$ \\
	$^{1}$University of Sydney \quad $^{2}$City University of Hong Kong \\
	\scriptsize\texttt{\{yuheng.shi, xiaohuan.pei, c.xu\}@sydney.edu.au, minjdong@cityu.edu.hk}
}

\newcommand{\methodname}{Vision-RL$^2$}
\iclrfinalcopy
\begin{document}

	\maketitle

	\begin{abstract}
	Fine-grained visual perception in multimodal large language models (MLLMs) is commonly improved by raising the resolution, but the added visual tokens inflate vision-encoding and language-model prefilling costs.
    We show that the two operations underlying fine-grained perception, \emph{localizing} the region of interest (RoI) and \emph{recognizing} its content, have different resolution requirements. In a controlled diagnostic, localization tolerates roughly $3$--$4\times$ stronger token compression than recognition, which motivates localizing from a coarse view and concentrating resolution on the selected evidence. 
    Decoding coordinates with the MLLM can be trained end-to-end from answers, but costs a full model pass per query and depends on grounding ability. A lightweight proposal network distilled from the model's attention is fast, but inherits the noise of its attention targets.
	The RoI from the proposal network reaches the answer through a discrete region choice, so its faithfulness to the answer cannot supervise the network. We therefore optimize the proposal network with region-level reinforcement learning, which we call \methodname{}. It treats coherent regions as actions, and a frozen MLLM reader scores each one by how its removal changes the answer likelihood.
	Complementary subtractive and additive objectives suppress distracting proposals and recover missing evidence, updating only the predictor without region annotations, response sampling, or reasoning trajectories. The refined proposal further enables a sparse encoding that magnifies evidence and excludes background tokens.
	Across six fine-grained benchmarks and four MLLM backbones, \methodname{} improves accuracy over the base model at every token budget and surpasses its largest-budget accuracy with about $4\times$ fewer visual tokens. Under the same token limit, \methodname{} also outperforms previous state-of-the-art methods that fully finetune the base model, while training an order of magnitude fewer parameters. Code is available at \url{https://github.com/YuHengsss/VisionRL2}.
\end{abstract}
    
	\section{Introduction}
\label{sec:intro}

Multimodal large language models (MLLMs)~\citep{Qwen2.5-VL,liu2023llava} still struggle with fine-grained details, as small text, distant objects, and cluttered high-resolution scenes reduce accuracy even when coarse scene understanding remains reliable~\citep{tong2024eyes,vstar,zhang2025mllms}.
The direct remedy is to increase the input image resolution, which produces more visual tokens and increases both vision-encoding and language-model costs.
As answer-relevant evidence often occupies a small fraction of the image, efficient perception requires allocating high-resolution processing selectively rather than uniformly across the image.

Answering a fine-grained visual question generally involves two perceptual operations, first \emph{localizing} the relevant evidence and then \emph{recognizing} its content.
Standard MLLMs and recent privileged-view distillation methods~\citep{wei2026zooming,yuan2026visionopd} optimize fine-grained perception within the full model, without exposing localization and recognition as separate stages.
Prior work shows quantitatively that MLLMs can concentrate attention on the ground-truth region even when they answer incorrectly~\citep{zhang2025mllms}, suggesting that recognition may fail despite an intact localization signal.
Following this insight, we build a controlled diagnostic on ZoomBench~\citep{wei2026zooming} with Qwen3.5-4B. The model predicts an RoI box from the scene, then answers from the resulting crop under a cap of at most $128$ visual tokens.
On filtered cases that this pipeline answers correctly at full resolution and that are not solvable from the question alone, the \emph{localization} sweep compresses only the scene from which the box is re-predicted, while the \emph{recognition} sweep freezes the box and compresses its crop. \emph{Survival} is the fraction of these cases still answered correctly.
At matched survival, localization tolerates roughly $3$--$4\times$ stronger token compression than recognition (Fig.~\ref{fig:teaser}a,b). The full protocol is given in Appendix~\ref{app:probe}.

\begin{figure}[t]
	\centering
	\vspace{-4mm}
	\includegraphics[width=0.99\textwidth]{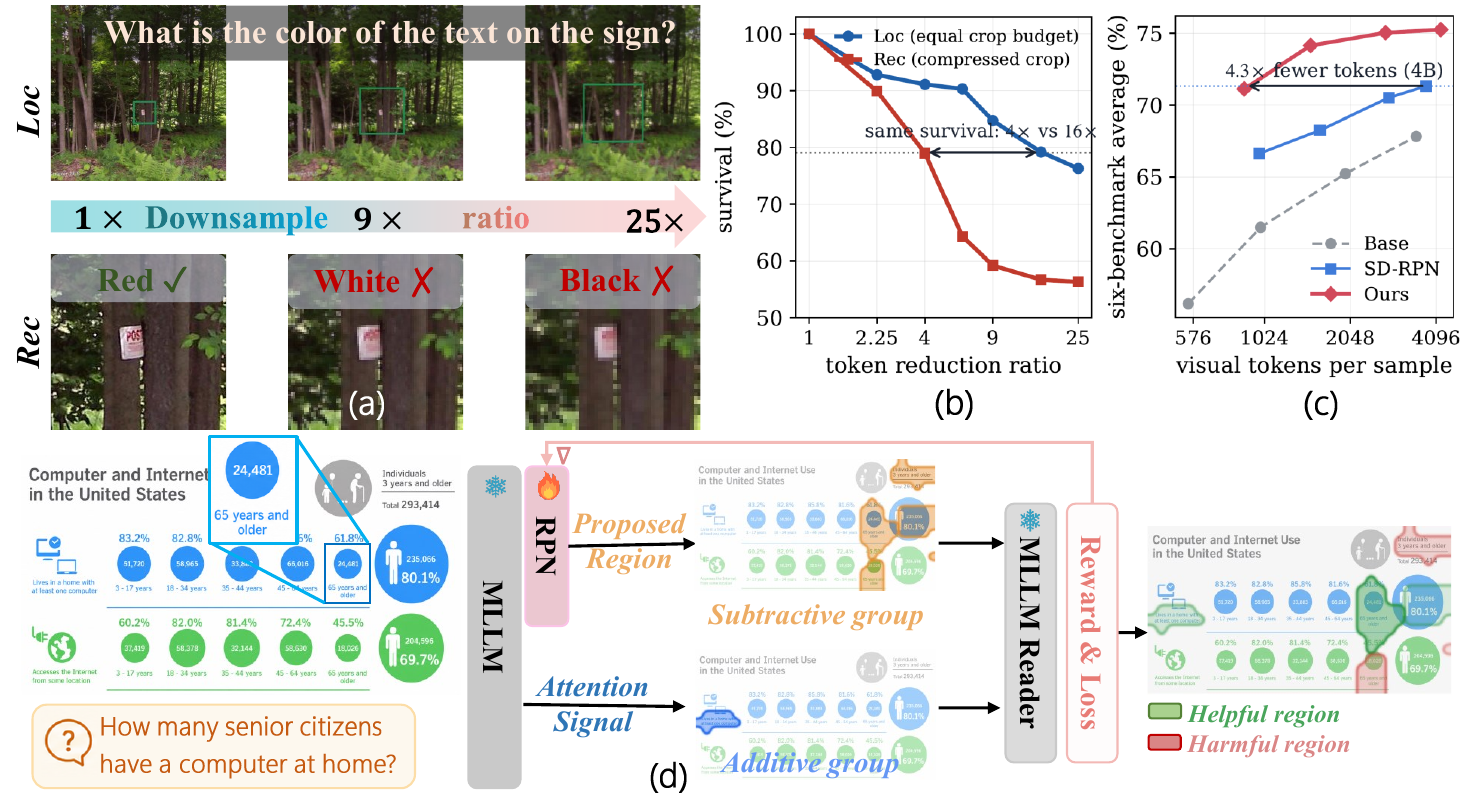}
	\vspace{-2mm}
	\caption{
		(a) The box is still re-predicted correctly at $25\times$ token reduction, while reading the crop fails at $9\times$.
		(b) Population survival ($n{=}238$) when compressing the locator's scene or the recognizer's crop in isolation. Recognition degrades more rapidly than localization.
		(c) Our method matches the SD-RPN's accuracy using $4.2\times$ fewer visual tokens on Qwen3.5-4B.
		(d) Overview of \methodname{}. The RPN proposes candidate regions and the MLLM supplies an attention-derived candidate. A reader scores every region's contribution to the answer and the reward updates the RPN.}
	\label{fig:teaser}
	\vspace{-6mm}
\end{figure}

This measured gap provides a quantitative basis for \emph{resolution-decoupled inference}, which localizes from a coarse view and reserves higher resolution for the selected evidence.
It places the burden on the RoI predictor, which must be reliable from a coarse view and add little overhead.
Existing two-stage systems implement such routing through attention-guided or iterative visual search~\citep{zhang2025mllms,shen2025zoomeye}, or through autoregressive coordinate generation~\citep{lee2026ergo,li2026p2r}.
Although effective, these interfaces require costly full-model operations, multiple interaction rounds, or explicit localization sequences.
Their routing overhead can therefore yield a less favorable accuracy--latency trade-off than privileged-view distillation methods.
SD-RPN~\citep{shi2026sdrpn} provides an efficient routing interface. A proposal network attached to intermediate MLLM layers predicts an answer-free RoI map in a single pass, trained on token-wise pseudo-labels distilled from response-to-image attention.
This surrogate objective may retain spurious activations or omit weakly attended evidence, and neither outcome is checked against the answer. Fig.~\ref{fig:teaser}d shows such a case, where the proposal keeps a distracting region and misses the evidence the answer needs.
Unlike a decoded box, which is part of the model's response and can be optimized from the answer directly, the RoI map reaches the answer only through a discrete region choice. The map is binarized, regions are extracted, and the selected crop is re-encoded, so no gradient connects the answer back to the map. Direct geometric supervision is also insufficient because the RoI in fine-grained VQA is not uniquely annotated and may be an arbitrary question-relevant region rather than a well-defined object~\citep{man2025argus}. What the predictor lacks is therefore an answer-level signal delivered with region-level credit, that is, a per-region measure of whether keeping it helped or harmed the answer.

We introduce \methodname{}, a \textbf{R}egion-\textbf{L}evel \textbf{R}einforcement \textbf{L}earning method that supplies exactly this signal (Fig.~\ref{fig:teaser}d).
Built on the trained SD-RPN predictor to retain its single-pass efficiency, \methodname{} treats coherent visual regions as actions, and a frozen MLLM reader scores each region by how its removal changes the teacher-forced likelihood of the gold answer.
A subtractive objective prunes predictions whose contribution falls below a calibrated noise margin, and an additive objective recovers evidence the proposal missed.
We further propose sparse visual encoding that processes the selected evidence at finer granularity at the token level.

Across multiple fine-grained benchmarks and backbones, \methodname{} outperforms both the base model and SD-RPN at every shared token budget (Fig.~\ref{fig:teaser}c).
Under the full-resolution setting, it is further competitive with recent state-of-the-art methods that fully finetune the MLLM, while updating only the proposal network.
In summary, we make three contributions. First, a controlled resolution intervention isolates localization and recognition and reveals their asymmetric compression tolerance. Second, our region-level policy optimization operates natively on a dense RoI map and trains a decode-free predictor from a frozen reader's functional answer signal without region annotations. Third, the learned predictor and sparse evidence encoding improve fine-grained accuracy and efficiency across models and benchmarks.

	\section{Related Work}
\label{sec:related}

MLLMs improve fine-grained perception primarily by admitting more visual tokens, through tiled high-resolution encoding~\citep{liu2024llavanext,chen2024internvl,li2024llava} or native dynamic-resolution encoders~\citep{Qwen2-VL,lu2025ovis2,dehghani2023patch,zhu2025internvl3}. Uniform scaling, however, spends encoding and prefilling compute regardless of where the evidence lies.
Beyond scaling, thinking-with-images methods optimize the full model with reinforcement learning to interleave decoding with crop and zoom actions~\citep{openai2025thinking,zheng2025deepeyes,lai2025mini,yang2025visionthink,zhang2025thyme,lee2026ergo,li2026p2r}, and latent-reasoning variants compress this search into the visual latent space~\citep{wang2025monet,li2025latent}.
These methods obtain answer-aware behavior, but full-model RL is memory-intensive and unstable, and inference pays for decoded trajectories or repeated generation rounds at every query.
Recent analysis further suggests that their gains are dominated by improvements in the RL-optimized model itself rather than by the interleaved tool use~\citep{ma2026what, wei2026zooming}.
Privileged-view distillation instead trains the backbone offline with region-enhanced teachers~\citep{wei2026zooming,yuan2026visionopd}, transferring an answer-level signal at the cost of full-model finetuning and without exposing a reusable localizer.
Another line separates localization from recognition at inference time.
Training-free systems guide the zoom with handcrafted rules over internal attention or search trees~\citep{zhang2025mllms,zhong2025focus,liu2025hide}. They need no training but incur multiple prefilling or decoding passes, and their heuristics transfer poorly across models and scenes.
Supervised alternatives predict RoIs from curated coordinate annotations~\citep{jiang2025token,shi2025scaling}, at heavy data-curation and finetuning cost.
SD-RPN~\citep{shi2026sdrpn,shi2026q} removes both requirements by self-distilling response-to-image attention into a lightweight, answer-free RoI predictor over intermediate MLLM features.
Yet its token-wise surrogate supervision never verifies that the proposed regions support the answer.

	\vspace{-2mm}\section{Method}

\vspace{-1.5mm}\subsection{Preliminaries}

\textbf{Response-to-image attention.}
Given an image $I$ and a question $q$, an MLLM maps the image to $N_v$ visual embeddings $\mH_v^0=\mathcal{P}(\mathcal{E}_v(I))$ through vision encoder $\mathcal{E}_v$ and projector $\mathcal{P}$, then generates a response $y=(y_1,\ldots,y_T)$.
At layer $l$, with response-token queries $\mQ_y^l\in\mathbb{R}^{T\times d}$ and visual-token keys $\mK_v^l\in\mathbb{R}^{N_v\times d}$, the response-to-image attention and its spatial map are:
\begin{equation}
	\mA_{y\rightarrow v}^{l}
	=\operatorname{softmax}\!\left(
	\mQ_y^l(\mK_v^l)^{\top}/\sqrt{d}\right),
	\qquad
	\mM^{l}=\mathcal{G}\!\left(
	\frac{1}{T}\sum_{t=1}^{T}
	\mA_{y\rightarrow v}^{l}[t,:]\right).
	\label{eq:response_image_map}
\end{equation}
where $\mathcal{G}:\mathbb{R}^{N_v}\rightarrow\mathbb{R}^{H_g\times W_g}$ reshapes the $N_v=H_gW_g$ tokens onto the visual-token grid.
$\mM^l$ often highlights answer-relevant evidence, but it typically becomes available only after decoding.

\textbf{SD-RPN.}
We build on the Self-Distilled Region Proposal Network (SD-RPN)~\citep{shi2026sdrpn}, which distills this latent grounding signal into an efficient, answer-free predictor. It reuses the first $B$ frozen MLLM blocks as its backbone, attaches a small stack of trainable blocks, and predicts a dense RoI logit map from the last prefilling token, $\mZ_{\theta}
=\mathcal{G}\!\left(\mQ_{\mathrm{RoI}}\mK_{\mathrm{RoI}}^{\top}\right)$,
where $\mQ_{\mathrm{RoI}}$ and $\mK_{\mathrm{RoI}}$ are normalized linear projections of the last prefilling token's hidden state and of the visual features refined by the trainable blocks.
Their trainable parameters are denoted by $\theta$, and $\mZ_\theta,\mP_\theta\in\mathbb{R}^{H_g\times W_g}$ with $\mP_\theta=\sigma(\mZ_\theta)$, where $\sigma$ is the element-wise sigmoid.
SD-RPN is trained by self-distilling Eq.~\ref{eq:response_image_map}, where denoised response-attention maps label only high-confidence foreground and background tokens and ambiguous tokens are ignored.
This token-wise supervision remains a local surrogate. Spurious regions may survive, incomplete attention may omit necessary evidence, and ignored cells receive no task-level signal.
Our region-level RL instead optimizes what this surrogate cannot see. The predictor should retain a \emph{minimal sufficient support}, exactly the evidence the answer requires, keeping every functionally necessary region and recovering evidence the proposal missed.

\begin{figure}[t]
	\centering
	\includegraphics[width=0.99\textwidth]{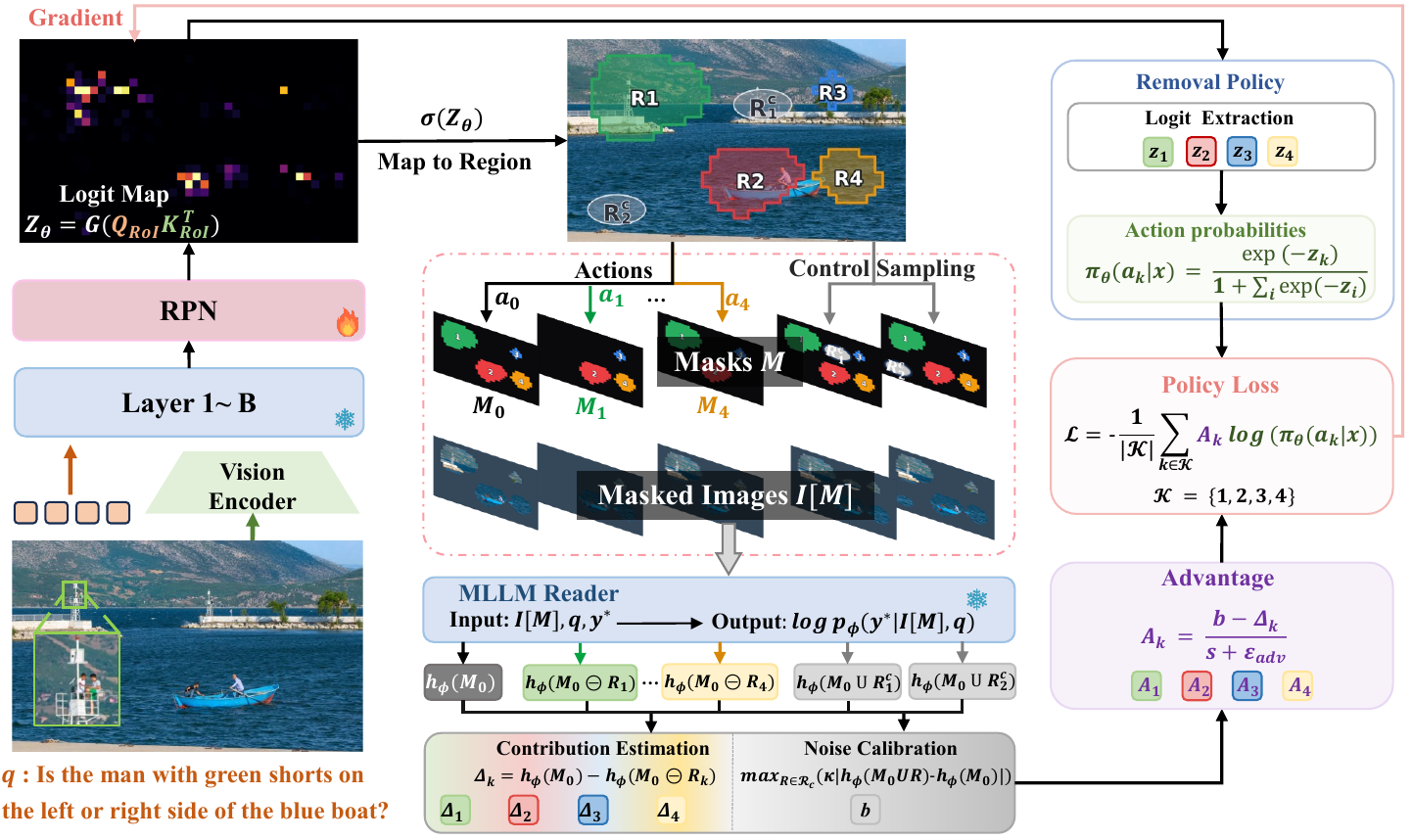}
	\caption{\textbf{Overview of the proposed region-level RL.} Predicted regions $R_1$--$R_4$ and control regions $R^{\mathrm{c}}_1,R^{\mathrm{c}}_2$ are extracted from the RoI map, the frozen reader scores each action's masked image, and the resulting contributions $\Delta_k$ and margin $b$ form the advantages and policy loss.
		Group markers are dropped for clarity, so $R_k,\Delta_k,A_k,s,\pi_\theta$ denote $R^{\mathrm{p}}_k,\Delta^{\mathrm{p}}_k,A^{\mathrm{p}}_k,s_{\mathrm{p}},\pi^{\mathrm{sub}}_\theta$ of Sec.~\ref{sec:functional_region_rl}, with $M_0=M_{\mathrm{p}}$ and $M_k=M_0\ominus R_k$. The additive group is omitted here for brevity.}
	\label{fig:method_overview}
	\vspace{-5mm}
\end{figure}

\vspace{-1.5mm}\subsection{Regions, Masks, and Functional Contributions}
\label{sec:region_action}
To realize this goal, we build three ingredients, namely regions as atomic units, masks that render retained regions into reader inputs, and each region's functional contribution to the answer.
Unlike a language model with fixed vocabulary, an RPN produces a dense heatmap rather than a distribution over RoI candidates. Treating the $H_gW_g$ cells as independent binary actions yields $2^{H_gW_g}$ masks and credits isolated cells, whereas cropping requires spatially coherent evidence and an answer-based signal is informative only for sufficiently complete supports.
We therefore take coherent visual regions as the primary objects of the formulation.
A \emph{region} $R$ is a connected set of visual-grid cells, produced from a spatial source map by one shared map-to-region operator. The map is smoothed, binarized at a peak-relative threshold, and split into connected components (detailed in Appendix~\ref{app:region_construction}).
Two sources instantiate this operator. The first is the current policy map $\mP_\theta$, and the second, used during training only, is a frozen response-to-image map of the form of Eq.~\ref{eq:response_image_map}.

Each region carries the grid indicator mask $m_R(u)=\mathbf{1}[u\in R]$, and a set $\mathcal{A}$ of retained regions is rendered by the mask constructor, $M(\mathcal{A})=\bigvee_{R\in\mathcal{A}} m_R$,
the cell-wise union of its members' masks.
A grid mask $M$ is upsampled to the image resolution, and $I[M]$ denotes the masked image, in which pixels outside the mask are replaced by the per-channel image mean (Fig.~\ref{fig:method_overview}). The reader below consumes only $I[M]$.

Rather than scoring geometry, we score function. A mask $M$ is good exactly when the evidence it retains still supports the correct answer.
Given the gold answer $y^\star=(y_1^\star,\ldots,y_T^\star)$, a frozen reader $p_\phi$, instantiated by the underlying MLLM itself, is conditioned on $I[M]$ and teacher-forced on the answer tokens. The geometric mean answer probability and its log-odds are:
\begin{equation}
	P_\phi(M)
	=\exp\!\left(\frac{1}{T}\sum_{t=1}^{T}
	\log p_\phi\!\left(
	y_t^\star\mid I[M],q,y_{<t}^\star\right)\right),
	\qquad
	h_\phi(M)=\operatorname{logit}(P_\phi(M)).
	\label{eq:functional_likelihood}
\end{equation}
Since $I$, $q$, and $y^\star$ are fixed per sample, $h_\phi(M)$ is a function of the mask alone, high when $M$ preserves the evidence required for $y^\star$ and dropping when that evidence is masked out.
We use it as the functional score. Its additive, unbounded scale supports relative comparisons between masks, scoring needs neither response sampling nor a reasoning trajectory, and it is treated as a constant (stop-gradient) during optimization.
The bridge from a single region to the functional score is its leave-one-out contribution against a reference mask $M_{\mathrm{ref}}$:
\begin{equation}
	\Delta_\phi(R\mid M_{\mathrm{ref}})
	=h_\phi(M_{\mathrm{ref}})-h_\phi(M_{\mathrm{ref}}\ominus R),
	\label{eq:generic_region_contribution}
\end{equation}
where $M\ominus R=M\odot(1-m_R)$ removes the cells of $R$ from a mask.
A positive contribution indicates that removing $R$ reduces functional support. Contributions are clipped to $[-\delta,\delta]$ with $\delta=5$.
The estimator is agnostic to how regions are produced and needs only $n+1$ reader evaluations for $n$ regions (one reference mask and $n$ singleton leave-one-out masks), rather than a search over subsets. The two action groups of Sec.~\ref{sec:functional_region_rl} instantiate it with different region sources and masks.

\vspace{-1.5mm}\subsection{Functional Region-Level Policy Optimization}
\label{sec:functional_region_rl}
Searching for this minimal sufficient support over candidate masks directly is impractical. Their number grows combinatorially with the number of regions, and a proposal that omits necessary evidence cannot be repaired by pruning alone.
We therefore approximate it with two marginal action groups, both instantiating Eq.~\ref{eq:generic_region_contribution}. The subtractive group removes predicted regions whose contribution does not exceed the noise level, and the additive group recovers unpredicted regions with positive contribution (Fig.~\ref{fig:method_overview} shows one step).

\textbf{Subtractive policy with noise calibration.}
The first group operates on the policy's own predictions.
Applying the shared map-to-region operator of Sec.~\ref{sec:region_action} to the policy map $\mP_\theta$ yields candidate components.
We rank the candidates by their mean RoI logit $z_\theta(R)=\frac{1}{|R|}\sum_{u\in R}\mZ_\theta(u)$ and keep the top $K\leq K_{\max}$ as the \textbf{p}olicy-predicted regions $\mathcal{R}_{\mathrm{p}}=\{R^{\mathrm{p}}_1,\ldots,R^{\mathrm{p}}_K\}$, with mask $M_{\mathrm{p}}=M(\mathcal{R}_{\mathrm{p}})$.
Region extraction is non-differentiable, but $z_\theta(R)$, the region's confidence, remains differentiable in the logits it averages.
Against the intact prediction, each region's contribution is $\Delta^{\mathrm{p}}_k=\Delta_\phi(R^{\mathrm{p}}_k\mid M_{\mathrm{p}})$.

Small likelihood changes arise even when a removal does not alter answer-relevant evidence, and their scale varies across samples and reader sizes, so $\Delta^{\mathrm{p}}_k$ alone does not tell whether a change exceeds the reader's noise under masking.
We estimate a sample-specific margin from a small set $\mathcal{R}_{\mathrm{c}}$ of \textbf{c}ontrol regions (gray in Fig.~\ref{fig:method_overview}), grown outside the dilated prediction in areas with $\mP_\theta<0.02$ and matched in area to the median predicted region:
\begin{equation}
	b=\min\!\Bigl(
	\kappa\max_{R\in\mathcal{R}_{\mathrm{c}}}
	\bigl|h_\phi\!\left(M(\mathcal{R}_{\mathrm{p}}\cup\{R\})\right)
	-h_\phi(M_{\mathrm{p}})\bigr|,
	b_{\max}\Bigr),
	\label{eq:null_margin}
\end{equation}
where $\kappa$ controls the calibration strength and $b_{\max}=1$ caps the margin. We set $|\mathcal{R}_{\mathrm{c}}|=2$ here.

The region confidence $z_\theta$ routes the resulting credit into the dense map.
Denote the intact action by $a_0$, the singleton removal of $R^{\mathrm{p}}_k$ by $a_k$, write $z_k=z_\theta(R^{\mathrm{p}}_k)$ and $x=(I,q)$, and let $\mathcal{K}$ index the removals that keep the foreground non-empty (emptying the mask would measure the loss of all visual support rather than a region's marginal effect).
Since $z_k$ is the mean logit of the region's per-cell Bernoulli inclusion, the removal $a_k$ scores $-z_k$ while the intact action scores $0$, and a softmax over these scores gives the removal policy:
\begin{equation}
	\pi_\theta^{\mathrm{sub}}(a_k\mid x)
	=\frac{\exp(-z_k)}
	{1+\sum_{j\in\mathcal{K}}\exp(-z_j)},
	\qquad k\in\mathcal{K},
	\label{eq:removal_policy}
\end{equation}
where the unit term is the intact action.
All removals in $\mathcal{K}$ are evaluated, so $\pi_\theta^{\mathrm{sub}}$ is never sampled. It is the differentiable channel that assigns each action's credit to its region.
The margin-calibrated advantage and policy loss are:
\begin{equation}
	A^{\mathrm{p}}_k
	=\frac{b-\Delta^{\mathrm{p}}_k}{s_{\mathrm{p}}+\varepsilon_{\mathrm{adv}}},
	\qquad
	\mathcal{L}_{\mathrm{sub}}
	=-\frac{1}{|\mathcal{K}|}
	\sum_{k\in\mathcal{K}}
	A^{\mathrm{p}}_k\log
	\pi_\theta^{\mathrm{sub}}(a_k\mid x),
	\label{eq:group1_loss}
\end{equation}
where $s_{\mathrm{p}}$ is the standard deviation of $\{\Delta^{\mathrm{p}}_k\}_{k\in\mathcal{K}}$ and $\varepsilon_{\mathrm{adv}}=1$.
If $\Delta^{\mathrm{p}}_k>b$, removing $R^{\mathrm{p}}_k$ receives negative credit and the region is preserved by increasing $z_k$. Otherwise its confidence is decreased.
With $K=1$ no relative removal exists, so $\mathcal{L}_{\mathrm{sub}}=0$ and the anchor below handles the region.

\textbf{Additive policy.}
Removal alone cannot recover evidence the policy misses, so the second group tests \textbf{s}upplementary regions, marked by $\mathrm{s}$, proposed by the training-only source.
Frozen response-to-image maps of the form of Eq.~\ref{eq:response_image_map}, taken from six MLLM layers spread across depth (Appendix~\ref{app:region_construction}), are each passed through the shared region operator.
Residual regions outside $M_{\mathrm{p}}$ are then merged across layers, and the top merged candidates form $\mathcal{R}_{\mathrm{s}}=\{R^{\mathrm{s}}_1,\ldots,R^{\mathrm{s}}_J\}$ with $J\leq4$.
With the augmented reference mask $M_{\mathrm{aug}}=M(\mathcal{R}_{\mathrm{p}}\cup\mathcal{R}_{\mathrm{s}})$, each supplementary region's contribution is $\Delta^{\mathrm{s}}_j=\Delta_\phi(R^{\mathrm{s}}_j\mid M_{\mathrm{aug}})$.
No control-region margin is applied here. For a candidate with no true contribution, exclusion is already the zero-gradient default of the loss below, so a margin would only bias against weak genuine recoveries.
Credit reaches supplementary regions through the mean inclusion log-likelihood $\ell^{+}_\theta(R)=\frac{1}{|R|}\sum_{u\in R}\log\mP_\theta(u)$:
\begin{equation}
	A^{\mathrm{s}}_j
	=\frac{\Delta^{\mathrm{s}}_j}{s_{\mathrm{s}}+\varepsilon_{\mathrm{adv}}},
	\qquad
	\mathcal{L}_{\mathrm{add}}
	=-\sum_{j=1}^{J}
	A^{\mathrm{s}}_j\,\ell^{+}_\theta(R^{\mathrm{s}}_j),
	\label{eq:group2_loss}
\end{equation}
where $s_{\mathrm{s}}$ is the standard deviation of $\{\Delta^{\mathrm{s}}_j\}_{j=1}^{J}$.
Positive contribution $\Delta^{\mathrm{s}}_j$ raises a missing region, while negative contribution suppresses it.
The two credit channels match their decision structures. The subtractive group resolves a categorical choice among mutually exclusive removals, so its removals compete under the normalizer of Eq.~\ref{eq:removal_policy}, whereas each supplementary candidate is an independent inclusion decision judged on its own contribution, which a shared normalizer would distort.

\textbf{Stabilization and overall objective.}
Within-group normalization can amplify noise when the reader assigns low likelihood to the gold answer under every tested mask, so both policy losses are scaled by a detached attainability weight $w_i$, an EMA-normalized, clipped ratio of the sample's best attained gold likelihood (Appendix~\ref{app:stabilization}).
A frozen copy of the initial SD-RPN provides a KL anchor $\mathcal{L}_{\mathrm{KL}}$, and when $K_i=1$ a binary cross-entropy term $\mathcal{L}_{\mathrm{K1}}$ preserves the single-component proposal, both defined in Appendix~\ref{app:anchors}.
The per-sample objective is:
\begin{equation}
	\mathcal{L}_i
	=\mathcal{L}_{\mathrm{KL},i}
	+w_i\left(
	\mathcal{L}_{\mathrm{sub},i}
	+\mathcal{L}_{\mathrm{add},i}\right)
	+\mathbf{1}[K_i=1]\,
	\mathcal{L}_{\mathrm{K1},i}.
	\label{eq:complete_region_rl}
\end{equation}
Training updates only the RPN parameters $\theta$ with the MLLM frozen. At inference, a single answer-free RoI prediction from $(I,q)$ is derived as in SD-RPN.

\vspace{-1.5mm}\subsection{Sparse Visual Encoding}
\label{sec:sparse_encoding}

\begin{wrapfigure}{r}{0.52\textwidth}
	\vspace{-3mm}
	\centering
	\includegraphics[width=0.51\textwidth]{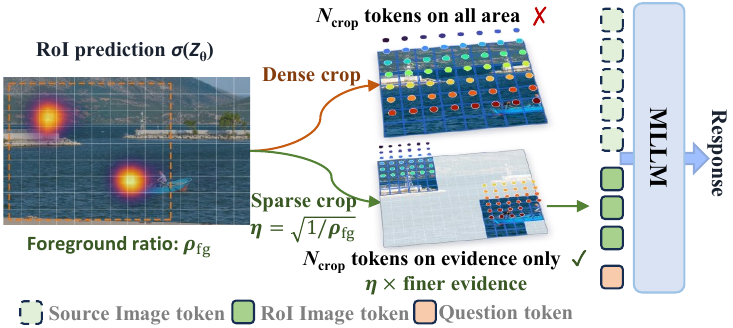}
	\caption{Sparse visual encoding.
		A dense crop spends the budget $N_{\mathrm{crop}}$ uniformly over the bbox, whereas the sparse crop encodes only foreground tokens at $\eta\times$ finer resolution.
		Source-image tokens are not re-encoded, and their KV cache is reused in part of the LLM layers.}
	\label{fig:sparse_encoding}
	\vspace{-4mm}
\end{wrapfigure}
At inference, the trained RoI predictor enables a simple principle: \emph{spend the visual-token budget on evidence, not on area} (Fig.~\ref{fig:sparse_encoding}).
Dense pipelines allocate tokens uniformly over the image or a crop, so background inside the crop dilutes the effective resolution of the evidence.
Given the RoI prediction, we take the bounding box of the predicted foreground, measure its foreground occupancy $\rho_{\mathrm{fg}}\in(0,1]$, and re-encode the crop at spatial zoom $\eta=\min\!\left(\sqrt{1/\rho_{\mathrm{fg}}},\eta_{\max}\right)$,
retaining only the visual tokens inside the foreground while background tokens never enter the vision encoder or the language model.
Position embeddings in both the vision encoder and the LLM are assigned on the full bbox-crop grid before background tokens are dropped, so the sparse token set is positionally indistinguishable from a dense encoding of the same crop.
Since the crop area grows by $\eta^2$, the retained-token count $\rho_{\mathrm{fg}}\eta^2N_{\mathrm{crop}}$ approximately matches the dense crop budget $N_{\mathrm{crop}}$, while the evidence is viewed at $\eta\times$ finer spatial resolution.
When the prediction contains multiple regions, we crop their single enclosing bounding box rather than each region independently. Independent crops would be equally token-efficient, but each carries its own coordinate frame, weakening the spatial relations between regions that the shared crop grid encodes through position embeddings.
The source-image tokens are not re-encoded in the second stage, and their KV cache from the proposal pass is reused in part of the LLM layers. Since RoI prediction needs neither the gold answer nor a decoded response, inference adds only one lightweight proposal pass before decoding.

	\definecolor{oursblue}{RGB}{219,236,244}
\definecolor{groupgray}{gray}{0.88}
\definecolor{oursgray}{gray}{0.92}
\newcommand{\prelim}[1]{\textcolor{gray}{#1}}
\newcommand{\splitheader}[2]{\begin{tabular}{@{}c@{}}#1\\#2\end{tabular}}

\vspace{-2mm}\section{Experiments}
\label{sec:experiments}

\vspace{-1.5mm}\subsection{Experimental Setup}
\label{sec:exp_setup}

\noindent\textbf{Models and training.}
We validate our method on Qwen3.5 4B and 9B~\citep{Qwen3.5}, Qwen2.5-VL-7B~\citep{Qwen2.5-VL}, and the encoder-free Gemma-4-12B~\citep{team2026gemma} (Appendix~\ref{app:gemma}).
For each model, the RoI predictor follows the SD-RPN configuration, RL starts from the supervised SD-RPN checkpoint, and the full frozen MLLM serves as the reader $p_\phi$.
The RL pool contains 7K QA pairs from the VisualCoT training corpus~\citep{shao2024visual}, with 5K from its InfographicVQA~\citep{mathew2022infographicvqa} split and 1K each from its TextVQA~\citep{singh2019towards} and DocVQA~\citep{mathew2020docvqa} splits.
Within each split of 10K candidates, samples are ranked by the standard deviation of their region-removal rewards under the initial SD-RPN and drawn from the top half.
The source image is encoded under a $576$ visual-token limit during training, so all region actions and rewards operate in this regime.
All models are trained for one epoch with a single recipe (batch size 32 and learning rate $1.5\times10^{-5}$, full configuration in Appendix~\ref{app:hyperparams}).

\noindent\textbf{Benchmarks and evaluation protocols.}
We evaluate on the high-resolution fine-grained perception benchmarks V* Bench~\citep{vstar}, ZoomBench~\citep{wei2026zooming}, HR-Bench 4K and 8K~\citep{hrbench}, and MME-RealWorld~\citep{zhang2024mme}.
The \emph{main protocol} (Table~\ref{tab:main_results}) follows the evaluation setting of recent fine-grained perception works~\citep{wei2026zooming,yuan2026visionopd}, where all methods share a $16{,}384$ source-image token limit, produce free-form responses, and are scored by rule-based parsing with an LLM judge for the remaining cases.
The \emph{training-aligned protocol}, used for the ablations and token-budget analyses, instead evaluates under the $576$-token training limit, extended to $\{576,1024, 2048,4096\}$ for the token-budget curves, with short-answer prompting, direct responses, and purely rule-based scoring, keeping the ablated components in the regime the objective optimizes (per-benchmark prompts in Appendix~\ref{app:eval_protocols}).

\vspace{-1.5mm}\subsection{Main Results}
\label{sec:main_results}

\begin{table}[t]
	\centering
	\caption{Comparison with SoTA MLLMs under a shared $16{,}384$ source-image token limit. Rows are grouped by base-model family, with large-scale open/closed-source models as reference.}
	\label{tab:main_results}
	\small\renewcommand{\arraystretch}{0.89}
	\setlength{\tabcolsep}{5.5pt}
	\resizebox{\textwidth}{!}{%
		\begin{tabular}{l c cccccc c}
			\toprule
			\textbf{Model} & \textbf{Size} & \textbf{V* Bench} & \textbf{ZoomBench} & \splitheader{\textbf{HR-Bench}}{\textbf{4K}} & \splitheader{\textbf{HR-Bench}}{\textbf{8K}} & \splitheader{\textbf{MME-RW}}{\textbf{EN}} & \splitheader{\textbf{MME-RW}}{\textbf{CN}} & \textbf{Average} \\
			\midrule
			\rowcolor{groupgray}
			\multicolumn{9}{c}{\textbf{Large-Scale Open/Closed-Source Models}} \\
			GPT-5.4 & --- & 77.0 & 52.7 & 84.0 & 77.9 & 74.2 & 70.9 & 72.8 \\
			Gemini-3.1-Pro & --- & 88.0 & 61.2 & 89.6 & 86.9 & 76.5 & 73.3 & 79.3 \\
			Qwen3-VL-Instruct & 235B & 91.1 & 56.1 & 86.1 & 80.4 & 71.7 & 69.0 & 75.8 \\
			Qwen3.5 & 397B & 88.0 & 57.2 & 89.4 & 85.5 & 74.8 & 69.8 & 77.4 \\
			Kimi-K2.6 & 1T & 88.5 & 53.1 & 81.9 & 78.0 & 69.2 & 66.1 & 72.8 \\
			\midrule
			\rowcolor{groupgray}
			\multicolumn{9}{c}{\textbf{Qwen2.5-VL Based}} \\
			DeepEyes & 7B & 85.9 & 46.5 & 75.1 & 72.6 & 64.1 & 64.1 & 68.1 \\
			Thyme & 7B & 82.2 & 45.1 & 77.0 & 72.0 & 64.8 & 64.6 & 67.6 \\
			DeepEyesV2 & 7B & 81.7 & 45.0 & 77.9 & 73.8 & 64.9 & 65.1 & 68.0 \\
			ZwZ & 7B & 86.9 & 55.6 & 75.9 & 72.4 & 65.0 & 63.5 & 69.9 \\
			\rowcolor{oursblue}
			\methodname{} (Ours) & 7B & 91.6 & 59.8 & 78.8 & 75.0 & 62.2 & 58.7 & 71.0 \\
			\midrule
			\rowcolor{groupgray}
			\multicolumn{9}{c}{\textbf{Qwen3-VL Based}} \\
			Qwen3-VL-Instruct & 8B & 84.8 & 43.0 & 79.6 & 75.3 & 63.2 & 64.6 & 68.4 \\
			ZwZ & 8B & 90.6 & 58.0 & 84.4 & 81.6 & 69.9 & 69.2 & 75.6 \\
			P2R & 4B & 93.2 & -- & 81.9 & 80.5 & -- & -- & -- \\
			P2R & 8B & 93.7 & -- & 81.5 & 82.6 & -- & -- & -- \\
			\midrule
			\rowcolor{groupgray}
			\multicolumn{9}{c}{\textbf{Qwen3.5 Based}} \\
			Qwen3.5 & 4B & 85.9 & 51.5 & 83.6 & 80.1 & 59.1 & 60.6 & 70.1 \\
			Qwen3.5 & 9B & 83.8 & 54.9 & 84.9 & 83.5 & 72.5 & 67.9 & 74.6 \\
			Vision-OPD & 4B & 90.6 & 59.5 & 82.0 & 79.1 & 74.2 & 70.6 & 76.0 \\
			Vision-OPD & 9B & 90.6 & 65.1 & 87.1 & 85.6 & 73.2 & 70.3 & 78.7 \\
			\rowcolor{oursblue}
			\methodname{} (Ours) & 4B & 91.1 & 65.1 & 84.3 & 80.3 & 65.8 & 65.3 & 75.3 \\
			\rowcolor{oursblue}
			\methodname{} (Ours) & 9B & 95.3 & 68.4 & 86.8 & 86.1 & 73.4 & 70.6 & 80.1 \\
			\midrule
			\rowcolor{groupgray}
			\multicolumn{9}{c}{\textbf{Gemma-4 Based}} \\
			Gemma-4 & 12B & 72.8 & 46.5 & 75.5 & 67.5 & 65.2 & 52.4 & 63.3 \\
			SD-RPN & 12B & 78.0 & 57.0 & 82.4 & 77.5 & 65.7 & 53.1 & 69.0 \\
			\rowcolor{oursblue}
			\methodname{} (Ours) & 12B & 82.2 & 60.7 & 85.0 & 79.4 & 67.6 & 61.6 & 72.8 \\
			\bottomrule
	\end{tabular}}
	\vspace{-5mm}
\end{table}

Table~\ref{tab:main_results} compares our models with SoTA fine-grained perception methods. Base-model and Vision-OPD/ZwZ rows are re-evaluated by us on released weights, the large-scale reference models~\citep{team2025kimi,Qwen3.5} are quoted from Vision-OPD~\citep{yuan2026visionopd}, and the rest from their publications~\citep{zhang2025thyme,hong2025deepeyesv2}.
A key distinction is the trainable parameter budget. DeepEyes~\citep{zheng2025deepeyes}, ZwZ~\citep{wei2026zooming}, P2R~\citep{li2026p2r}, and Vision-OPD~\citep{yuan2026visionopd} fine-tune the full MLLM, whereas our method updates only the small attached predictor.
Despite this, the 9B model attains the highest average in the table, ahead of Gemini-3.1-Pro and Vision-OPD-9B, with the best V* Bench and ZoomBench scores overall and a lead over Vision-OPD-9B on every remaining benchmark except HR-Bench 4K.
At the 4B scale our model surpasses Vision-OPD-4B on V* Bench, ZoomBench, and both HR-Bench splits, while trailing on the MME-RealWorld splits, where full fine-tuning can adapt the reader itself, a gap our frozen-reader design deliberately forgoes.
On the Qwen2.5-VL backbone, our 7B model attains the highest average of its block, ahead of ZwZ-7B's $69.9$, and leads it on every benchmark except the MME-RealWorld splits, the same trade as the smaller Qwen3.5 model. Gemma-4-12B has no native-resolution mode, so its rows use the largest visual-token tier ($1{,}120$ tokens), where our predictor adds $3.8$ points over SD-RPN and $9.5$ over the base model on average. Qualitative comparisons are given in Appendix~\ref{sec:Qualitative_comparison}.

\vspace{-1.5mm}\subsection{Efficiency Analysis}
\label{sec:efficiency}

\begin{figure}[t]
	\centering
	\includegraphics[width=0.95\textwidth]{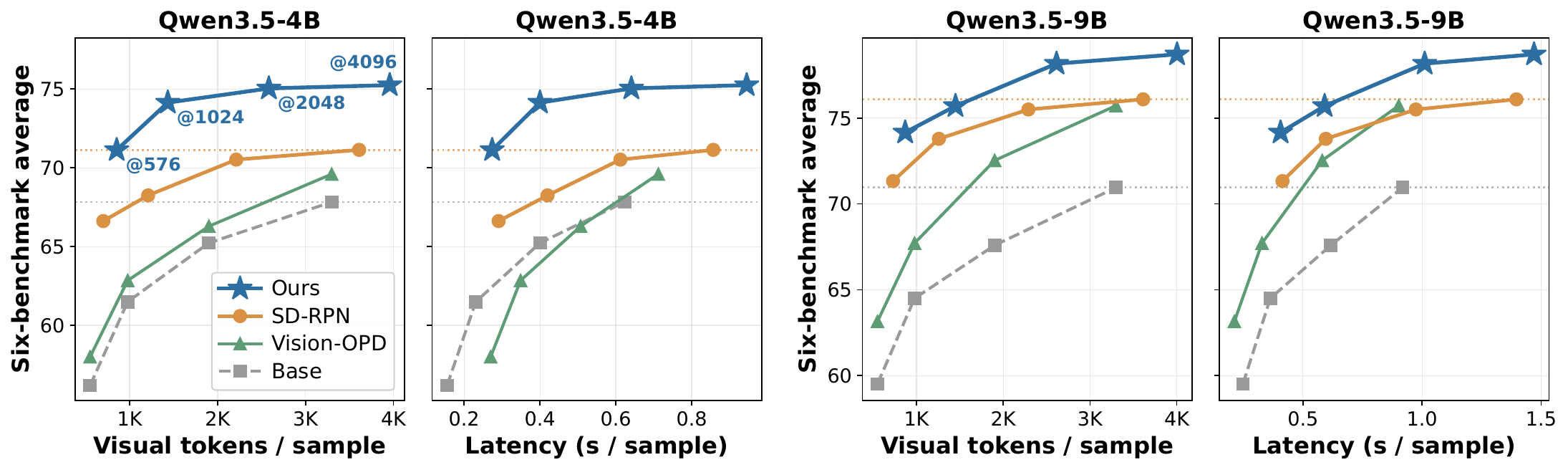}
	\caption{Six-benchmark average accuracy versus inference cost under the training-aligned short-answer protocol, for source-image token limits $\{576, 1024, 2048, 4096\}$. For each model scale, the left panel measures cost in visual tokens per sample (source plus crop) and the right panel as end-to-end latency. Per-benchmark token curves are given in Appendix~\ref{app:perbench_curves}.}
	\label{fig:tradeoff}
	\vspace{-7mm}
\end{figure}
\begin{wrapfigure}{r}{0.52\linewidth}
	\vspace{-7mm}
	\centering
	\includegraphics[width=\linewidth]{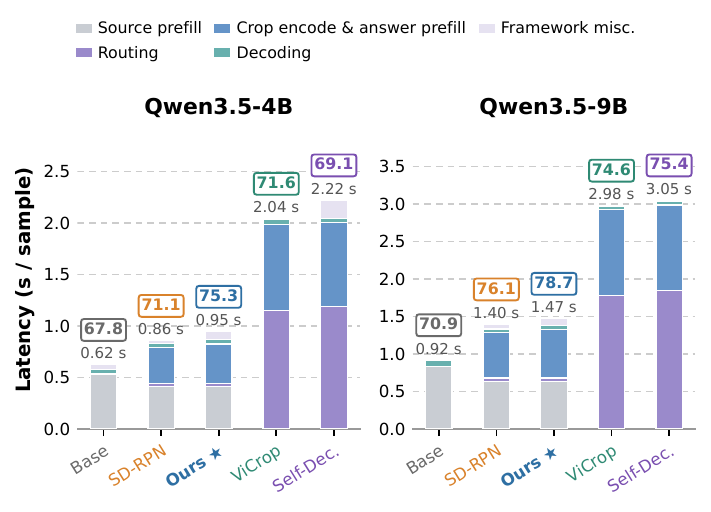}
	\caption{Latency decomposition of two-stage routing mechanisms at the $4{,}096$-token source limit, with the six-benchmark average accuracy above each bar. ViCrop~\citep{zhang2025mllms} and Self-Dec. incur substantial cost for RoI routing. All methods share the protocol and GPU.}
	\label{fig:routing_bars}
	\vspace{-5mm}
\end{wrapfigure}
All values here are produced under one shared protocol with lmms-eval~\citep{zhang2024lmmsevalrealitycheckevaluation} on the same RTX A6000 GPU.
	To keep the analysis tractable, the large MME-RealWorld EN/CN splits are replaced by MME-RealWorld-Lite and the InfoVQA validation split, and models answer directly rather than free-form, following the two-stage inference convention of SD-RPN~\citep{shi2026sdrpn}. Values are therefore not comparable with the main table.

Fig.~\ref{fig:tradeoff} compares the base model, SD-RPN, Vision-OPD, and ours over source-image token limits. Our model lies above all baselines at every budget on all six benchmarks (per-benchmark curves in Appendix~\ref{app:perbench_curves}).
At both scales, our model under the $576$ limit exceeds the base model at the $4{,}096$ limit by more than three points while consuming about one quarter of its tokens.
Our model also matches the SD-RPN accuracy at the $4{,}096$ limit with $4.2\times$ fewer visual tokens on the 4B model and reaches within half a point with $2.5\times$ fewer on the 9B model.
In wall-clock latency (right panels), our curve dominates the accuracy--latency plane at both scales. The 4B model at the $1{,}024$ limit exceeds SD-RPN at the $4{,}096$ limit with $2.1\times$ lower latency, the 9B model reaches within one point of that level at $2.4\times$ lower latency, and both models at the $576$ limit surpass the base model at the $4{,}096$ limit at $2.3\times$ lower latency.
Against the released Vision-OPD weights, our 4B model at the $576$ limit exceeds Vision-OPD-4B at its largest budget while responding $2.6\times$ faster.

Figure~\ref{fig:routing_bars} compares the three routing interfaces of Sec.~\ref{sec:intro} under one setting. Attention routing~\citep{zhang2025mllms} and coordinate decoding (the MLLM itself decodes the box) spend a full source pass, plus box tokens for the latter, purely on localization, taking $1.1$--$1.8$\,s per query, or $2.1$--$2.2\times$ our end-to-end latency. The RPN head adds three blocks on the answer call's own prefill and routes in $31$--$49$\,ms, roughly $30\times$ cheaper, while reaching the best accuracy at both scales.

\vspace{-1.5mm}\subsection{Ablation Studies}
\label{sec:ablation}

All ablations use the training-aligned protocol of Sec.~\ref{sec:exp_setup} on Qwen3.5-4B with all remaining recipe constants fixed, and the sparse visual encoding is used only where marked.
Tables report the six-benchmark average of the efficiency analysis, which adds InfoVQA~\citep{mathew2022infographicvqa} and differs from the main-table average.
The top block of Table~\ref{tab:mechanism_ablation} isolates the two mechanisms added on top of the supervised predictor. Region-level RL alone contributes $+3.0$ average over SD-RPN, and the sparse visual encoding adds a further $+1.5$ at the same source-token limit.
The gains are complementary, with RL improving which evidence is proposed and the sparse encoding how the fixed crop budget is spent on it. Relative to the frozen base model, the full system gains $+14.9$.

\begin{table}[t]
	\centering
	\caption{Ablations on Qwen3.5-4B under the training-aligned protocol ($576$-token source limit). Top block, mechanism ablation, where R.L.\ denotes region-level RL and S.E.\ the sparse visual encoding of Sec.~\ref{sec:sparse_encoding}. Bottom block, RL design ablation grouped by design axis with the sparse encoding disabled. $\Delta$ is the change in six-benchmark average vs.\ the full recipe.}
	\label{tab:mechanism_ablation}
	\label{tab:rl_ablation}
	\small\renewcommand{\arraystretch}{0.86}
	\setlength{\tabcolsep}{4pt}
	\resizebox{\textwidth}{!}{%
		\begin{tabular}{@{}l cccccc c c@{}}
			\toprule
			\textbf{Variant} & \textbf{V* Bench} & \textbf{ZoomBench} & \splitheader{\textbf{HR-Bench}}{\textbf{4K}} & \splitheader{\textbf{HR-Bench}}{\textbf{8K}} & \splitheader{\textbf{MME-RW}}{\textbf{Lite}} & \textbf{InfoVQA} & \textbf{Average} & $\Delta$ \\
			\midrule
			\prelim{Qwen3.5-4B (base)} & \prelim{66.0} & \prelim{40.5} & \prelim{63.5} & \prelim{56.4} & \prelim{41.0} & \prelim{69.8} & \prelim{56.2} & \prelim{---} \\
			\midrule
			\multicolumn{9}{@{}l}{\emph{Mechanism:}} \\
			\quad SD-RPN & 82.7 & 55.6 & 71.1 & 63.3 & 48.9 & 78.1 & 66.6 & $-3.0$ \\
			\rowcolor{oursgray}
			\quad SD-RPN + R.L.\ (full recipe) & 82.2 & 58.7 & 77.3 & 69.1 & 50.4 & 80.1 & 69.6 & --- \\
			\quad SD-RPN + R.L.\ + S.E. & \textbf{85.3} & \textbf{61.8} & \textbf{77.4} & \textbf{70.9} & \textbf{51.0} & \textbf{80.5} & \textbf{71.1} & $+1.5$ \\
			\midrule
			\multicolumn{9}{@{}l}{\emph{Reward: functional score $\rightarrow$}} \\
			\quad Generation accuracy & 80.6 & 53.3 & 73.9 & 65.1 & 50.7 & 80.5 & 67.3 & $-2.3$ \\
			\quad Raw log-probability & 81.7 & 53.1 & 75.8 & 67.6 & 50.0 & 80.4 & 68.1 & $-1.5$ \\
			\midrule
			\multicolumn{9}{@{}l}{\emph{Actions: singleton region removals $\rightarrow$}} \\
			\quad Singleton + pair removals & 83.3 & 56.7 & 76.8 & 69.8 & 51.4 & 80.5 & 69.7 & $+0.1$ \\
			\quad Cell-level keep-sets & 84.3 & 57.0 & 72.9 & 65.6 & 50.6 & 80.2 & 68.4 & $-1.2$ \\
			\midrule
			\multicolumn{9}{@{}l}{\emph{Credit assignment:}} \\
			\quad No control-region margin ($\kappa=0$) & 82.2 & 54.2 & 76.0 & 69.1 & 49.7 & 79.8 & 68.5 & $-1.1$ \\
			\quad No additive group & 82.2 & 55.6 & 75.5 & 68.4 & 50.1 & 80.3 & 68.7 & $-0.9$ \\
			\bottomrule
	\end{tabular}}
	\vspace{-6mm}
\end{table}

The lower block ablates the RL objective itself, grouped by design axis, with the sparse encoding disabled throughout.
On the reward axis, replacing the functional score with generation accuracy costs $-2.3$, since binary correctness gives no relative credit when all actions in a group share an outcome, and replacing the clipped log-odds score with the raw mean log-probability costs $-1.5$, since saturated samples then dominate the gradient scale.
On the action axis, adding pair removals (with a group-mean baseline, as the margin is calibrated for singletons) performs at parity with the full recipe at clear extra cost, so we keep the leave-one-out design.
A budget-matched cell-level control (not single-token pruning), which samples keep-sets of the predicted-foreground size cell-wise from the map logits and rewards them as complete masks, costs $-1.2$, confirming that spatially coherent regions are the right action unit.
On the credit axis, removing the control-region margin costs $-1.1$ and removing the additive recovery group costs $-0.9$, so an explicit noise floor is needed even with the intact-prediction reference, and subtractive-only training misses recoverable evidence.

	\vspace{-2mm}\section{Conclusion}
\label{sec:conclusion}

	We start from a measurement that localizing evidence tolerates several times stronger token compression than recognizing its content.
	\methodname{} turns this asymmetry into a training principle. A lightweight RoI predictor is made answer-accountable through region-level reinforcement learning, where a frozen MLLM reader scores each coherent region by its functional contribution to the gold answer, without region annotations or response sampling.
	Across fine-grained benchmarks and backbones, it outperforms both the base model and its supervised predecessor at every token budget while routing RoIs efficiently.
	The trained predictor is a reusable, decode-free localizer, and we expect its recipe of scoring proposals by their measured effect on a frozen reader to extend to other interfaces that decide where to spend computation.

	\bibliography{iclr2027_conference}
	\bibliographystyle{iclr2027_conference}
	\clearpage
	\appendix
	
\section{Method Details}
\label{app:method_details}

\subsection{Region Construction}
\label{app:region_construction}

Both candidate sources of the main text share one map-to-region operator.
For a spatial source map $\mM$:
\begin{equation}
	\widetilde{\mM}=G_{\sigma_g}*\mM,
	\qquad
	\mathfrak{R}(\mM)
	=\operatorname{CC}\!\left(
	\mathbf{1}\!\left[
	\widetilde{\mM}>
	\rho\max\nolimits_u\widetilde{\mM}(u)
	\right]\right),
	\label{eq:shared_regionization}
\end{equation}
where $G_{\sigma_g}$ is a Gaussian kernel, $\rho$ is a peak-relative threshold, and $\operatorname{CC}$ extracts connected regions.
Source maps whose peak-to-mean ratio indicates diffuse, non-discriminative activation are discarded.

For the subtractive source, $\mathfrak{R}$ is applied to the current policy map $\mP_\theta$. Region extraction and top-$K_{\max}$ ranking are non-differentiable, while the region confidences $z_\theta(R)$ remain differentiable with respect to the logits inside each retained region.
For the additive source, frozen response-to-image maps from layers $7$, $11$, $15$, $19$, $23$, and $27$ are each passed through $\mathfrak{R}$. The layers are spaced evenly across the depth of the MLLM because answer-relevant attention emerges at different depths for different samples, and no single layer covers all of them.
The predicted mask $M_{\mathrm{p}}$ is subtracted, overlapping residual regions are merged across layers, and the top merged candidates form $\mathcal{R}_{\mathrm{s}}$.

\subsection{Anchor Losses}
\label{app:anchors}

Let $\mP_{\theta_0}$ denote the RoI probability map of the frozen initial SD-RPN, computed at the same grid resolution as $\mP_\theta$.
The KL anchor is the mean per-cell Bernoulli divergence from the current map to this reference:
\begin{equation}
	\mathcal{L}_{\mathrm{KL}}
	=\frac{1}{H_gW_g}\sum_{u}
	D_{\mathrm{KL}}\!\left(
	\mathrm{Bern}(\mP_\theta(u))
	\,\big\|\,
	\mathrm{Bern}(\mP_{\theta_0}(u))\right),
	\label{eq:kl_anchor}
\end{equation}
applied to all samples so the map does not drift on prompts the policy gradient cannot reach.
When the component set collapses to a single region ($K=1$), no valid relative removal exists and the policy loss is zero. The sample is instead regularized by a binary cross-entropy toward the binarized reference foreground:
\begin{equation}
	\mathcal{L}_{\mathrm{K1}}
	=\mathrm{BCE}\!\left(
	\mP_\theta,\;
	\mathbf{1}\!\left[\mP_{\theta_0}>0.5\right]\right),
	\label{eq:k1_anchor}
\end{equation}
which preserves the single-component proposal in that regime.

\subsection{Stabilization Details}
\label{app:stabilization}

\paragraph{Attainability weight.}
Both policy losses are scaled by a detached attainability weight that suppresses samples on which the frozen reader cannot attain the gold answer under any tested intervention:
\begin{equation}
	P_{\max}^{(i)}
	=\max_{M\in\mathcal{M}_{\mathrm{sub}}^{(i)}}P_\phi(M),
	\qquad
	w_i=\operatorname{clip}\!\left(
	\frac{P_{\max}^{(i)}}
	{\max(\operatorname{EMA}_{0.99}[P_{\max}],0.05)},
	0,3\right),
	\label{eq:attainability_weight}
\end{equation}
where $\mathcal{M}_{\mathrm{sub}}^{(i)}$ collects the subtractive-group masks of sample $i$, namely the intact prediction $M_{\mathrm{p}}$ and all retained singleton-removal masks.

\section{Experimental Details}
\label{app:exp_details}

\subsection{Hyperparameters}
\label{app:hyperparams}

Table~\ref{tab:hyperparams} lists the training configuration and the constants of the method section, all of which are shared across the four backbones.
Two settings are model-specific and stated here instead.
The RoI predictor attaches three trainable blocks after the first $B$ frozen MLLM blocks, with $B=21$ for the two Qwen3.5 models, $B=18$ for Qwen2.5-VL-7B, and $B=27$ for Gemma-4-12B, following each backbone's SD-RPN configuration~\citep{shi2026sdrpn}.
The control-margin scale is $\kappa=1.25$ for Qwen3.5-4B and $\kappa=1.0$ for Qwen3.5-9B, Qwen2.5-VL-7B, and Gemma-4-12B, whose source images are encoded at the $560$-token tier, the closest tier to the $576$-token limit.

\begin{table}[hbt!]
	\centering
	\caption{Training hyperparameters, shared across the four backbones.}
	\label{tab:hyperparams}
	\small
	\begin{tabular}{@{}ll@{}}
		\toprule
		\textbf{Config} & \textbf{Setting} \\
		\midrule
		Optimizer & AdamW \\
		Weight decay & 0.0 \\
		Optimizer momentum & $(\beta_1,\beta_2)=(0.9,0.999)$ \\
		Global batch size & 32 \\
		Learning rate schedule & cosine decay \\
		Peak learning rate & $1.5\times10^{-5}$ \\
		Warm-up strategy / ratio & linear / 0.2 \\
		Source-image token limit & 576 \\
		\midrule
		Max predicted regions $K_{\max}$ & 6 \\
		Max supplementary regions $J$ & 4 \\
		Margin cap $b_{\max}$ & 1.0 \\
		Contribution clip $\delta$ & 5.0 \\
		Advantage regularizer $\varepsilon_{\mathrm{adv}}$ & 1.0 \\
		\midrule
		Region-operator smoothing $\sigma_g$ / threshold $\rho$ & 1.0 / 0.3 \\
		Diffuse-map discard (peak-to-mean) & $<3.0$ \\
		\bottomrule
	\end{tabular}
\end{table}

\subsection{Evaluation Prompts}
\label{app:eval_protocols}

\newtcolorbox{promptbox}[1]{enhanced,colback=white,colframe=black,
	colbacktitle=black,coltitle=white,fonttitle=\bfseries,title=#1,
	boxrule=0.8pt,arc=2.5mm,left=2mm,right=2mm,top=1.5mm,bottom=1.5mm,
	fontupper=\ttfamily\small}

\paragraph{Prompts of benchmarks.}
For reproducibility, we list the complete prompts used under both evaluation protocols.
MME-RealWorld-CN uses the Chinese translation of the EN template, and MME-RealWorld-Lite reuses the same template under the training-aligned protocol with the trailing short-answer line.

\begin{promptbox}{V* Bench / HR-Bench (main protocol)}
<image>\\
\{question\} Select from the following choices.\\
\{options\}
\end{promptbox}

\begin{promptbox}{ZoomBench (main protocol)}
<image>\\
\{question\}\\
\{options\}
\end{promptbox}

\begin{promptbox}{MME-RealWorld EN (main protocol)}
<image>\\
\{question\} The choices are listed below:\\
\{options\}\\
Select the best answer to the above multiple-choice question based on the image. Respond with only the letter (A, B, C, D, or E) of the correct option.\\
The best answer is:
\end{promptbox}

\begin{promptbox}{V* Bench (training-aligned protocol)}
<image>\\
\{question\}\\
\{options\}\\
Answer with the option's letter from the given choices directly.
\end{promptbox}

\begin{promptbox}{HR-Bench (training-aligned protocol)}
<image>\\
\{question\}\\
\{options\}\\
Answer the option letter directly.
\end{promptbox}

\begin{promptbox}{ZoomBench (training-aligned protocol)}
<image>\\
\{question\}\\
\{options\}\\
Answer with the option's letter from the given choices.
\end{promptbox}

\begin{promptbox}{InfoVQA (training-aligned protocol)}
<image>\\
\{question\}\\
Answer the question using a single word or phrase.
\end{promptbox}

\subsection{Per-Benchmark Token Curves}
\label{app:perbench_curves}

\begin{figure}[hbt!]
	\centering
	\includegraphics[width=\textwidth]{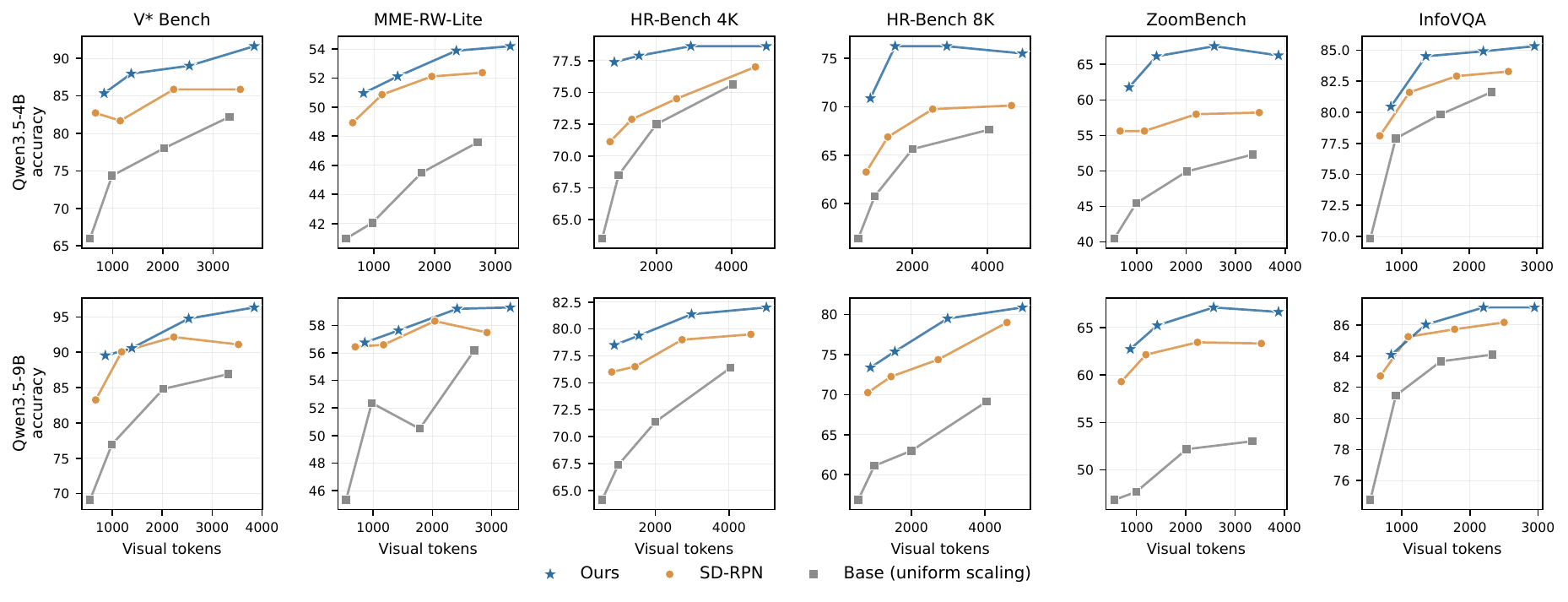}
	\caption{Per-benchmark accuracy versus measured visual tokens per sample (source plus crop) for source-image token limits $\{576, 1024, 2048, 4096\}$ under the training-aligned short-answer protocol (top row Qwen3.5-4B, bottom row Qwen3.5-9B).}
	\label{fig:token_curves_perbench}
\end{figure}

Figure~\ref{fig:token_curves_perbench} breaks the six-benchmark average of Figure~\ref{fig:tradeoff} down into per-benchmark accuracy-versus-token curves. Our model lies above both the base model and the SD-RPN pipeline at every source-image token limit on all six benchmarks and at both scales.

\subsection{Generalization to an Encoder-Free Backbone}
\label{app:gemma}

Gemma-4-12B~\citep{team2026gemma} has no vision transformer: images are cut into $48\times48$ patches that a linear embedder maps directly into the language model at one of five visual-token tiers ($70$ to $1{,}120$ tokens), and the processor always fills the chosen tier.
We follow the same pipeline as for the Qwen backbones with the tier as the only budget knob.
The SD-RPN predictor is trained with the Qwen3.5 recipe on the same corpus regenerated by Gemma-4 at the $560$-token tier, RL uses this predictor's own $7$K pool at the same tier, and the source image is encoded at the largest tier ($1{,}120$ tokens) at test time.
The RoI crop is encoded at the tier that quantizes the crop budget of Sec.~\ref{sec:sparse_encoding}, and sparse visual encoding drops background crop tokens before the embedder.

\subsection{Comparison with Visual Token Pruning}
\label{app:pruning}

\providecommand{\retc}[1]{{\scriptsize\textcolor{gray}{#1\%}}}

Visual token pruning reduces inference cost by selecting a subset of the encoded tokens, and is the main alternative route to visual-token efficiency. We compare against VisionZip~\citep{yang2024visionzip} and DART~\citep{wen2025dart}, the two methods with released Qwen2.5-VL implementations, on Qwen2.5-VL-7B. All methods run in the same evaluation harness with identical prompts, using each method's released code. Every method and benchmark uses one uniform pixel range, from $4$ to $4{,}096$ visual tokens at native aspect ratio.

\textbf{Token accounting.} The reference point is the base model under a $4{,}096$-token source limit ($100\%$). The pruning methods encode this full-budget source and then retain a fixed fraction of each sample's tokens. VisionZip keeps dominant-plus-contextual tokens after the vision encoder and DART prunes to the same nominal fraction at the second LLM layer, so their $50\%$ and $25\%$ budgets are exact by construction. Our method instead re-allocates resolution before encoding. On the high-resolution suite we report the $1{,}024$- and $576$-token rungs of the source ladder, whose measured totals (source plus RoI crop) average $41.2\%$ and $24.7\%$ of the reference. On the document suite, where native sizes vary widely, we downsample the source to $\mathrm{src}/a$ ($a{=}3,6$) and cap the crop at half the downsampled source, so the per-sample total is close to $1.5\,\mathrm{src}' \approx 50\%$ and $25\%$ of native, with measured means of $46.6\%$ and $27.2\%$.

\begin{figure}[hbt!]
	\centering
	\includegraphics[width=0.92\textwidth]{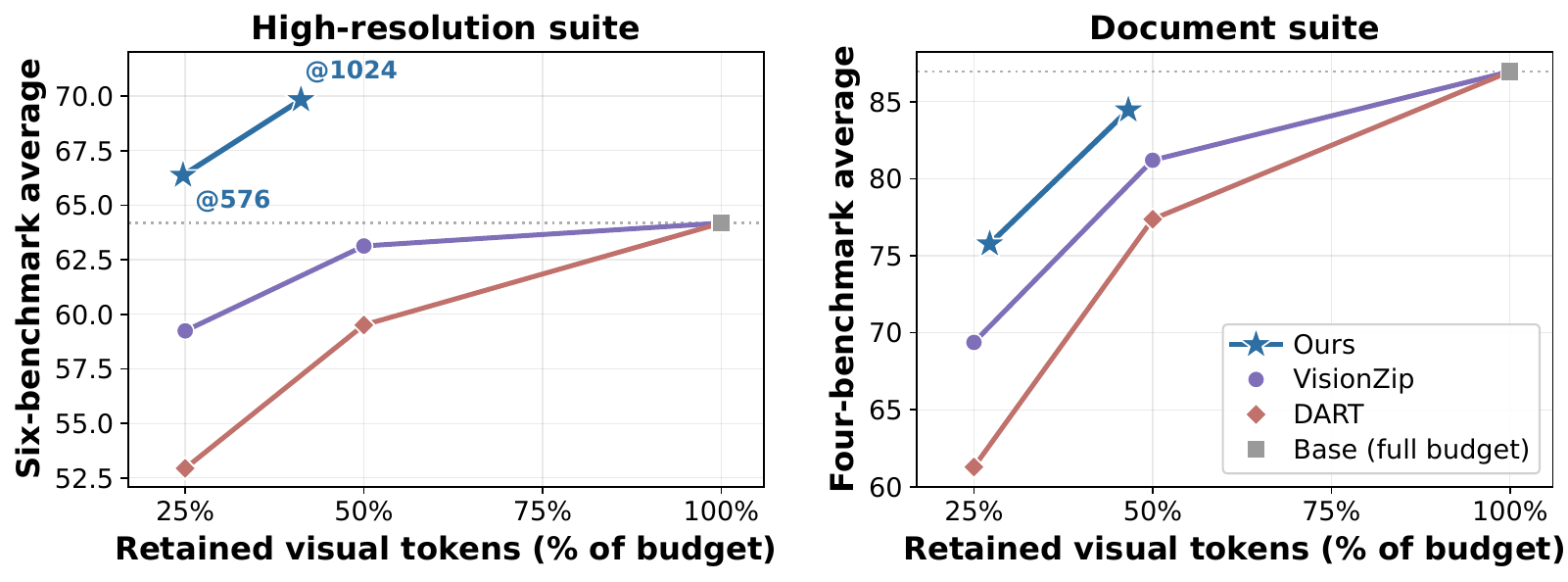}
	\caption{Suite-average accuracy versus retained visual-token budget for the pruning comparison of Table~\ref{tab:pruning_comparison}. Pruning points sit at their exact keep fractions and converge to the base model at $100\%$. Our points sit at their measured totals. On the high-resolution suite our curve lies above the full-budget reference at every budget.}
	\label{fig:pruning_trend}
\end{figure}

Table~\ref{tab:pruning_comparison} reports accuracy with per-benchmark retention relative to the full-budget reference, and Figure~\ref{fig:pruning_trend} plots the suite averages against the retained-token budget. At the $50\%$ tier, VisionZip is nearly lossless on the high-resolution suite, consistent with its published claims. Our model does not merely retain the reference accuracy but exceeds it at both tiers. On text-dense benchmarks pruning collapses. At the $25\%$ tier VisionZip retains $73.3\%$ on InfoVQA and $56.5\%$ on OCRBench, and DART falls further, while our method holds $94$--$99\%$ on both. The one cell where pruning wins is ChartQA at the $25\%$ tier, where chart understanding depends on the global layout, so at one-sixth source resolution uniform token selection preserves more structure than region re-allocation. Finally, the two approaches spend compute differently. Pruning still pays the full vision encoding and saves only LLM-side tokens, whereas our routing reduces both the encoder and LLM budgets.

\begin{table}[hbt!]
	\centering
	\caption{Comparison with visual token pruning on Qwen2.5-VL-7B under a shared $4{,}096$-token source limit. Gray subrows give accuracy retention relative to the full-budget base model. Pruning budgets are exact per-sample keep fractions. Ours are measured source-plus-crop totals, quoted per tier in the group headers.}
	\label{tab:pruning_comparison}
	\small
	\setlength{\tabcolsep}{4pt}
			\begin{tabular}{@{}l cccccc c@{}}
			\toprule
			\textbf{Method} & \textbf{V* Bench} & \textbf{ZoomBench} & \splitheader{\textbf{HR-Bench}}{\textbf{4K}} & \splitheader{\textbf{HR-Bench}}{\textbf{8K}} & \splitheader{\textbf{MME-RW}}{\textbf{Lite}} & \textbf{InfoVQA} & \textbf{Avg.\ ret.} \\
			\midrule
			\rowcolor{groupgray}
			\multicolumn{8}{c}{\emph{High-resolution suite}} \\
			Base & 77.49 & 46.86 & 72.50 & 64.00 & 42.31 & 81.89 & 100\% \\
			\rowcolor{groupgray}
			\multicolumn{8}{c}{\emph{Retain $50\%$ (ours measured at $41.2\%$)}} \\
			VisionZip & 79.06 & 44.62 & 72.00 & 63.75 & 43.04 & 76.28 & \multirow{2}{*}{98.4\%} \\
			& \retc{102.0} & \retc{95.2} & \retc{99.3} & \retc{99.6} & \retc{101.7} & \retc{93.1} & \\
			DART & 78.53 & 45.21 & 68.75 & 60.50 & 39.19 & 64.88 & \multirow{2}{*}{92.7\%} \\
			& \retc{101.3} & \retc{96.5} & \retc{94.8} & \retc{94.5} & \retc{92.6} & \retc{79.2} & \\
			\textbf{Ours} & 85.86 & 54.79 & 76.50 & 70.87 & 47.79 & 83.16 & \multirow{2}{*}{\textbf{108.8\%}} \\
			& \retc{110.8} & \retc{116.9} & \retc{105.5} & \retc{110.7} & \retc{113.0} & \retc{101.6} & \\
			\rowcolor{groupgray}
			\multicolumn{8}{c}{\emph{Retain $25\%$ (ours measured at $24.7\%$)}} \\
			VisionZip & 81.68 & 43.43 & 69.25 & 61.38 & 39.66 & 60.02 & \multirow{2}{*}{92.3\%} \\
			& \retc{105.4} & \retc{92.7} & \retc{95.5} & \retc{95.9} & \retc{93.7} & \retc{73.3} & \\
			DART & 73.82 & 42.37 & 65.00 & 58.00 & 33.82 & 44.61 & \multirow{2}{*}{82.5\%} \\
			& \retc{95.3} & \retc{90.4} & \retc{89.7} & \retc{90.6} & \retc{79.9} & \retc{54.5} & \\
			\textbf{Ours} & 80.10 & 51.95 & 72.00 & 66.38 & 46.59 & 81.18 & \multirow{2}{*}{\textbf{103.4\%}} \\
			& \retc{103.4} & \retc{110.9} & \retc{99.3} & \retc{103.7} & \retc{110.1} & \retc{99.1} & \\
			\bottomrule
		\end{tabular}

	\vspace{2mm}

			\begin{tabular}{@{}l cccc c@{}}
			\toprule
			\textbf{Method} & \textbf{OCRBench} & \textbf{ChartQA} & \textbf{DocVQA} & \textbf{TextVQA} & \textbf{Avg.\ ret.} \\
			\midrule
			\rowcolor{groupgray}
			\multicolumn{6}{c}{\emph{Document suite}} \\
			Base & 85.50 & 84.48 & 94.86 & 82.98 & 100\% \\
			\rowcolor{groupgray}
			\multicolumn{6}{c}{\emph{Retain $50\%$ (ours measured at $46.6\%$)}} \\
			VisionZip & 70.20 & 78.96 & 94.03 & 81.62 & \multirow{2}{*}{93.4\%} \\
			& \retc{82.1} & \retc{93.5} & \retc{99.1} & \retc{98.4} & \\
			DART & 70.10 & 70.84 & 87.96 & 80.57 & \multirow{2}{*}{89.0\%} \\
			& \retc{82.0} & \retc{83.9} & \retc{92.7} & \retc{97.1} & \\
			\textbf{Ours} & 80.60 & 79.92 & 94.92 & 82.38 & \multirow{2}{*}{\textbf{97.1\%}} \\
			& \retc{94.3} & \retc{94.6} & \retc{100.1} & \retc{99.3} & \\
			\rowcolor{groupgray}
			\multicolumn{6}{c}{\emph{Retain $25\%$ (ours measured at $27.2\%$)}} \\
			VisionZip & 48.30 & 66.80 & 86.91 & 75.46 & \multirow{2}{*}{79.8\%} \\
			& \retc{56.5} & \retc{79.1} & \retc{91.6} & \retc{90.9} & \\
			DART & 51.60 & 53.56 & 68.70 & 71.27 & \multirow{2}{*}{70.5\%} \\
			& \retc{60.4} & \retc{63.4} & \retc{72.4} & \retc{85.9} & \\
			\textbf{Ours} & 70.20 & 60.60 & 94.03 & 78.25 & \multirow{2}{*}{\textbf{87.1\%}} \\
			& \retc{82.1} & \retc{71.7} & \retc{99.1} & \retc{94.3} & \\
			\bottomrule
		\end{tabular}
\end{table}

\subsection{Qualitative Comparison}
\label{sec:Qualitative_comparison}

\begin{figure}[t]
	\centering
	\includegraphics[width=0.99\textwidth]{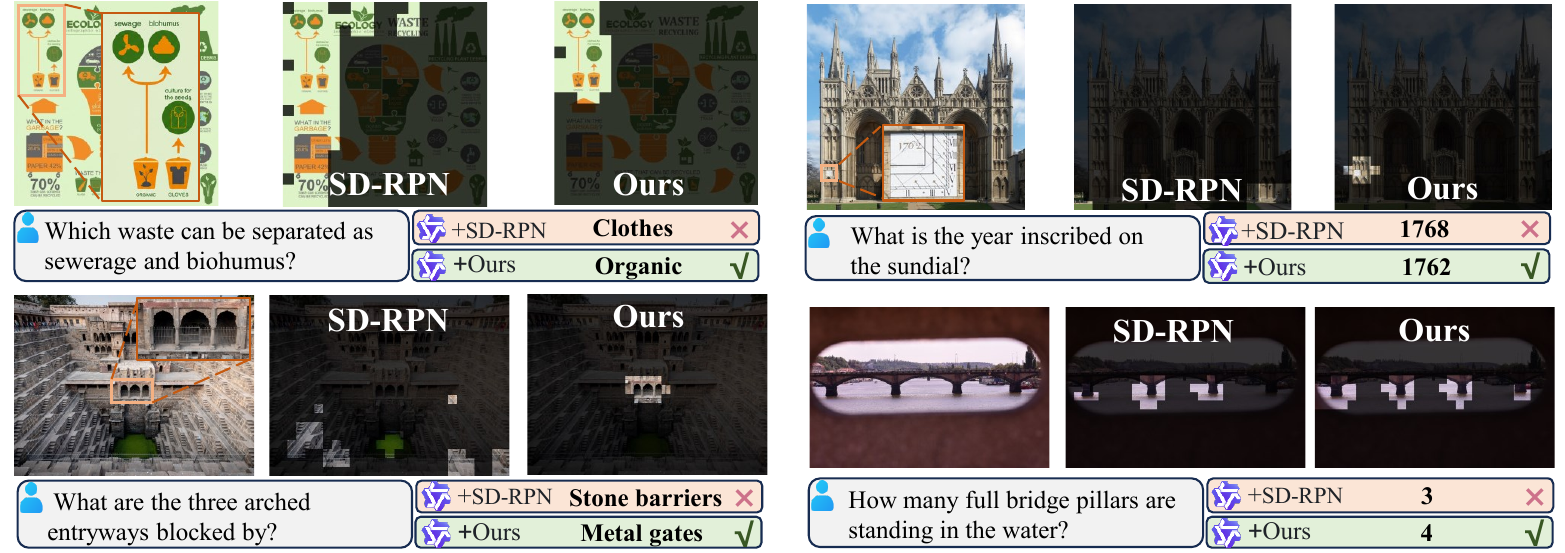}
	\caption{RoI predictions of SD-RPN and ours. Dimmed regions are the discarded tokens.}
	\label{fig:qualitative}
	\vspace{-5mm}
\end{figure}

 Figure~\ref{fig:qualitative} illustrates how the two predictors differ on cases where their answers diverge. The supervised SD-RPN exhibits the two failure modes of its pseudo-label surrogate. It scatters probability over question-irrelevant regions and covers the evidence only partially, so the reader is fed the wrong tokens and answers incorrectly. After region-level RL refinement, the same predictor suppresses the spurious regions and retains the required evidence, from the sundial plaque and the gated archways to all four bridge pillars, so the reader answers correctly at the same token budget. This is the intended effect of the objective. Rewards are assigned to regions by their functional contribution to the answer, so the predictor learns to keep exactly the evidence the answer depends on rather than whatever correlates with the teacher's attention.

\section{Localization-vs-Recognition Probe: Protocol and Controls}
\label{app:probe}

\paragraph{Setup.}
The probe uses the Qwen3.5-4B base model with no RoI predictor attached, on ZoomBench~\citep{wei2026zooming}, whose items are four-way multiple-choice and blank-answer questions. The mixed chance level of this pool is the shaded band in Fig.~\ref{fig:teaser}b.
The model itself serves as the locator. Given the question and an instruction to output the evidence box, it decodes a bounding box greedily, with the assistant reply prefilled by the JSON prefix \texttt{\{"bbox\_2d": [} so that every response is parseable.
We restrict the pool to samples that are answerable from the ground-truth-box crop, and remain so under a $15\%$ area-preserving jitter of that box, but are not confidently solvable from the question text alone under a first-token margin filter. This leaves $|E|{=}276$ samples and removes the text-prior contamination that would otherwise grow as image information vanishes.
The anchor run at $r{=}1$ presents the native-resolution scene. We record the predicted box and keep the $n{=}238$ eligible samples that emit a parseable box and are answered correctly from the model's own predicted RoI. Survival at compression $r$ is the probability of a correct answer at $r$ given a correct answer at the anchor, measured on this fixed set.

\paragraph{Ladders.}
The localization ladder downsamples the scene by $r$, so that its visual tokens scale as $1/r^2$ with no processor floor, re-predicts the box from the degraded scene, and reads the answer from the corresponding crop at native resolution.
The recognition ladder freezes the anchor box, excludes the background, and downsamples the crop itself by $r$ before encoding. The processor floors the crop at about six tokens from $r{\ge}3$ onward, so recognition survival at the deep rungs is, if anything, optimistic.

\paragraph{Fixed-crop-budget control.}
Along the localization ladder the re-predicted box grows with $r$, reaching a median area ratio of $7.4\times$ at $r{=}6$, and a larger box buys a larger native-resolution crop.
To rule out this budget effect, a control re-runs the ladder with the crop hard-capped at $128$ tokens at every rung (Table~\ref{tab:probe_control}). Even at a fixed crop budget, localization retains an $18.2$-point advantage over recognition at $r{=}6$.

\begin{table}[h]
	\centering
	\caption{Survival (\%) on the $n{=}238$ anchor-correct set under scene compression (localization, with and without the crop cap) and crop compression (recognition).}
	\small
	\begin{tabular}{lcccccccc}
		\toprule
		$r$ & 1.5 & 2 & 2.5 & 3 & 4 & 5 & 6 & 10 \\
		\midrule
		Localization (uncapped)      & 93.7 & 91.2 & 91.2 & 84.0 & 84.0 & 82.8 & 82.4 & 76.9 \\
		Localization (crop $\le$128 tok) & 92.8 & 91.1 & 90.3 & 84.7 & 79.2 & 76.3 & 69.5 & 55.0 \\
		Recognition                  & 89.9 & 79.0 & 64.3 & 59.2 & 56.7 & 56.3 & 51.3 & 44.5 \\
		\bottomrule
	\end{tabular}
	\label{tab:probe_control}
\end{table}

\paragraph{Box behavior.}
Along the localization ladder the re-predicted box loses precision but keeps containing the evidence. Across the rungs from $r{=}1.5$ to $r{=}6$, median IoU against the anchor box falls as $0.73/0.62/0.54/0.42/0.23/0.16/0.11$, while median ground-truth coverage rises from $0.44$ to $1.00$ and the hit rate, defined as coverage of at least $0.5$, rises from $47\%$ to $75\%$.
Localization survival therefore reflects a graceful soft failure, an imprecision that downstream cropping recovers, rather than precision robustness.

\section{Limitations}
\label{app:limitations}

Our design deliberately keeps the reader frozen, so it cannot repair recognition failures that finetuning the full model can, which is visible on MME-RealWorld at the smallest scale.
The compression-asymmetry probe is a diagnostic on one backbone and one benchmark rather than a general law, and training requires question--answer pairs to score contributions, though no region annotations.

\end{document}